\documentclass[10pt]{article} % For LaTeX2e
\usepackage[preprint]{tmlr}
\usepackage{graphicx}
\usepackage{makecell}
\usepackage{braket}
\usepackage{xspace}

\usepackage{amsmath,amsfonts,bm}

\def\eqref#1{equation~\ref{#1}}
\def\1{\bm{1}}

\def\vk{{\bm{k}}}

\def\vx{{\bm{x}}}

\def\mA{{\bm{A}}}

\def\mD{{\bm{D}}}

\def\mI{{\bm{I}}}

\def\mS{{\bm{S}}}
\def\mT{{\bm{T}}}
\def\mU{{\bm{U}}}

\def\mX{{\bm{X}}}

\DeclareMathAlphabet{\mathsfit}{\encodingdefault}{\sfdefault}{m}{sl}
\SetMathAlphabet{\mathsfit}{bold}{\encodingdefault}{\sfdefault}{bx}{n}

\def\gG{{\mathcal{G}}}

\def\emA{{A}}

\newcommand{\E}{\mathbb{E}}

\newcommand{\R}{\mathbb{R}}

\newcommand*\Laplace{\mathop{}\!\mathbin\bigtriangleup}

\usepackage{hyperref}
\usepackage{url}

\title{QDC: Quantum Diffusion Convolution Kernels on Graphs}

\author{\name Thomas Markovich \email tmarkovich@squareup.com \\
      \addr CashApp \\
      Cambridge, Massachusetts, USA}

\newcommand{\V}{\mathcal{V}}
\renewcommand{\E}{\mathcal{E}}
\renewcommand{\L}{\mathcal{L}}
\newcommand{\Q}{\mathcal{Q}}
\newcommand{\iu}{\mathrm{i}\mkern1mu}
\newcommand{\methodname}{QDC\xspace}
\newcommand{\multimethodname}{MultiScaleQDC\xspace}

\def\month{MM}  % Insert correct month for camera-ready version
\def\year{YYYY} % Insert correct year for camera-ready version
\def\openreview{\url{https://openreview.net/forum?id=XXXX}} % Insert correct link to OpenReview for camera-ready version

\begin{document}

\maketitle

\begin{abstract}
 Graph convolutional neural networks (GCNs) operate by aggregating messages over local neighborhoods given the prediction task under interest. Many GCNs can be understood as a form of generalized diffusion of input features on the graph, and significant work has been dedicated to improving predictive accuracy by altering the ways of message passing. In this work, we propose a new convolution kernel that effectively rewires the graph according to the occupation correlations of the vertices by trading on the generalized diffusion paradigm for the propagation of a quantum particle over the graph. We term this new convolution kernel the Quantum Diffusion Convolution (\methodname) operator. In addition, we introduce a multiscale variant that combines messages from the \methodname operator and the traditional combinatorial Laplacian. To understand our method, we explore the spectral dependence of homophily and the importance of quantum dynamics in the construction of a bandpass filter. Through these studies, as well as experiments on a range of datasets, we observe that \methodname improves predictive performance on the widely used benchmark datasets when compared to similar methods.
\end{abstract}

\section{Introduction}

Graphs are mathematical structures composed of vertices and edges, where the edges represent potentially complex relationships or interactions between the vertices. Graph structured data involves complex patterns and relationships that can not be captured by the traditional deep learning methods that focus on tabular data. As a result, graph structured data and graph machine learning models have become increasingly important in many fields, such as machine learning~\citep{wu2020comprehensive}, computer vision~\citep{krzywda2022graph}, and natural language processing~\citep{wu2023graph}. Indeed, graphs and graph structured data are ubiquitous in industrial applications ranging from fraud detection~\citep{liu2020alleviating,zhang2022efraudcom}, to routing~\citep{rusek2019unveiling, chen2022approach}, weather predictions~\citep{keisler2022forecasting, ma2022histgnn}, drug discovery~\citep{bongini2021molecular, han2021reliable, xiong2021graph}, and personalized recommendations~\citep{wu2022graph, gao2021graph}.

Graph neural networks (GNNs) are an increasingly popular modality for constructing graph machine learning models~\citep{zhou2020graph, wu2020comprehensive}. There are many variations of architectures for GNNs, but most GNNs can be thought of as having a function that, for a given vertex, aggregates information from its neighbors, and a second function which maps this aggregated information to the machine learning task under investigation, such as node classification, node regression, or link prediction. A simple but powerful model is a Graph Convolutional network (GCN), which extends the convolutional neural network (CNN) architecture to the graph domain by using a localized filter that aggregates information from neighboring nodes~\citep{zhang2019graph}. By sharing weights across different nodes, GCNs can learn representations that capture both the local and global structure of the graph. These models have shown remarkable success in a variety of tasks such as node classification~\citep{gcn, zhang2019graph}, graph classification\citep{xie2020graph}, community detection~\citep{jin2019graph, wang2021unsupervised}, and link prediction~\citep{chen2020multi,cai2019transgcn,zeb2022complex}. GCNs learn filters on the graph structure to be used at inference time. Early GCN development learned these filters in the spectral domain~\citep{bruna2013spectral}, but this requires the decomposition of large matrices. Due to the computational expense of these decompositions, spatial filters rose in popularity and have been the dominant paradigm. Significant effort has been dedicated to methodological improvements that make spatial convolutions more expressive~\citep{bodnar2021weisfeiler, bouritsas2022improving} and scalable~\citep{hamilton2017inductive, ying2018graph}. Further work has shown that it is possible to unify many of these models based on spatial convolutions as generalized graph diffusion~\citep{grand, blend, digl}, with considerable attention being focused on improving diffusion dynamics~\citep{elhag2022graph, di2022graph}.

We take a different approach and consider the question: ``Can we improve message passing in graph neural networks with a quantum mechanical message passing kernel?'' The answer comes in the form of a graph Laplacian preprocessing framework that is very flexible to embed into the architecture of many graph neural networks. This framework allows us to define \methodname, a Quantum Diffusion Kernel that rewires the graph according to the likelihood a quantum particle starting at one vertex will be observed at another. Unlike the heat equation, which tends to maximize entropy, the use of quantum dynamics seems to improve coherence through the presence of constructive and destructive interference. In summary, this paper's core contributions are

\begin{itemize}
  \item We propose \methodname, a quantum mechanically inspired diffusion kernel, a more powerful and general method for computing sparsified non-local transition matrices.
  \item We propose a novel multi-scale message passing paradigm that performs message passing using \methodname and the original combinatorial Laplacian in parallel. 
  \item We compare and evaluate \methodname and \multimethodname to a set of similar baselines on a range of node classification tasks.
  \item We analyze the spectral dependence of homophily in graph datasets and show that many heterophilous datasets are actually homophilous in filtered settings.
\end{itemize}

\section{Related Work}

Our method can be viewed as a technique for graph rewiring, because it changes the computation graph from the original adjacency matrix to a filtered one. Graph rewiring is a common technique in the literature for improving GNN performance by removing spurious connections. The SDRL method seeks to perform this rewiring by fixing instances of negative curvature on the graph~\citep{topping2021understanding}. +FA attempts to improve performance by adding a fully adjacent layer to message passing~\citep{alon2020bottleneck}. GDC is the most similar method to ours, but is based on the idea of heat, rather than quantum, diffusion which yields a low pass filter~\citep{digl}.

Our method can also be understood within the context of physics inspired graph neural networks. GraphHEAT proposes performing graph convolutions using a parameterized version of the heat kernel~\citep{xu2020graph}. The GRAND method recasts message passing as anisotropic diffusion on a graph, and provides a framework with which to unify many popular GNN architectures~\citep{grand, grand++}. BLEND pushes this perspective further to explore diffusion in non-euclidean domains~\citep{blend}. PDE-GCN looks further and seeks to combine diffusion with the wave equation to define new message passing frameworks~\citep{pde-gcn}. To our knowledge, ours is the first work that explores quantum dynamics as a message passing formalism.

Our method is closely related to the kernel signatures that have been explored in computer graphics. The Heat Kernel Signature was one of the first kernel signatures that was developed, and was developed by modeling the heat diffusion over a surface~\citep{sun2009concise, zobel2011generalized}. It was observed that the Heat Kernel Signature was sensitive to motions that might introduce a bottle neck, such as an articulation, which motivated the development of the Wave Kernel signature. The Wave Kernel Signature defines surface structural features by modeling the infinite time diffusion of quantum particles over the surface~\citep{wave_kernel_signature}. Building on the wave kernel signature, the average mixing kernel signature explores the finite time diffusion of a quantum particle over the surface~\citep{cosmo2020average}. The wave kernel trace method instead explores solutions of the acoustic wave equation using the edge-Laplacian~\citep{aziz2014graph}. These methods have all been used to develop graph features that would then be used to identify similar shapes, but we are instead using these kernels as convolution operators.

\section{Background}\label{sec:background}

\begin{figure}[t]
  \begin{center}
    \includegraphics[width=\textwidth]{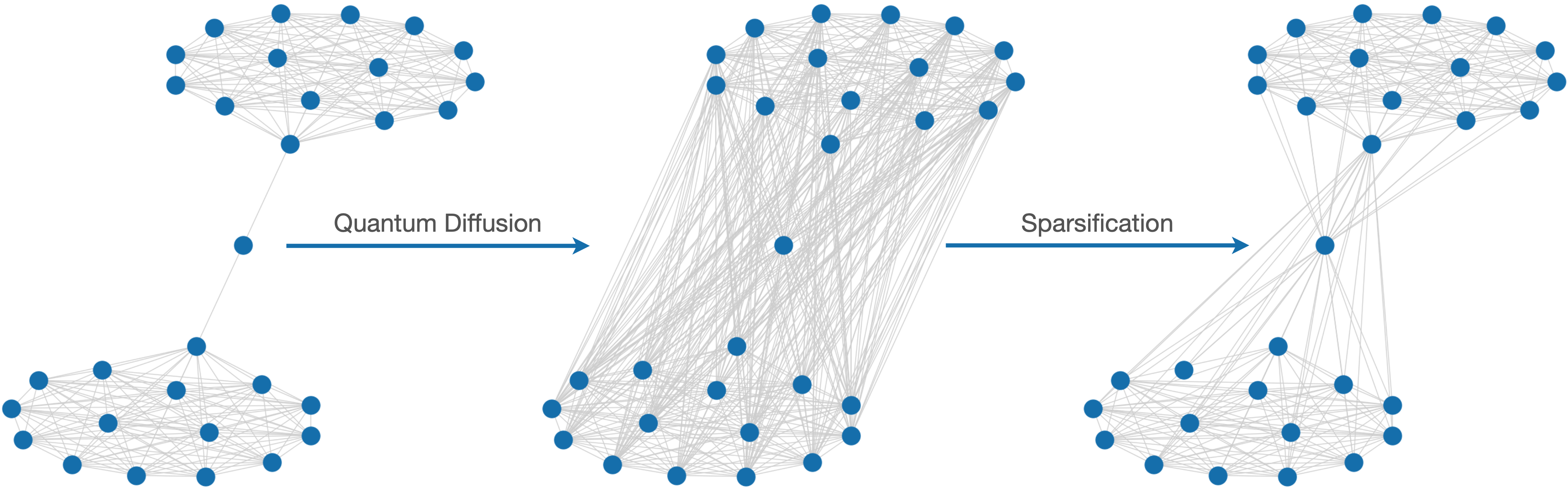}
  \end{center}
  \caption{An illustration of the Quantum Diffusion Convolution Kernel. The graph is transformed through the propagation of a quantum particle as given by \eqref{eq:qdc}, and then sparsified the graph to account for the presence of rare events. The model is then trained on the post-sparsified rewired graph.}
  \label{fig:fig1}
\end{figure}

\textbf{Preliminaries} Given an undirected graph, $\gG = (\V, \E, \mX)$, where $\V$ is the vertex set with cardinality $|\V| = N$, and $\E$ is the edge set, and $\mX \in \R^{N x d}$ denote the matrix of vertex features, where d is the dimensionality of the feature set. $\E$ admits an adjacency matrix, $\mA \in \R^{N x N}$, where $\emA_{ij} = 1$ if and only if vertices $i$ and $j$ are connected. Because we have restricted ourselves to undirected graphs, $\emA_{ij} = \emA_{ji}$. While this $\gG$ could have weighted edges, we focus on the unweighted case for simplicity. It is common to augment a graph with self loops, which is performed by $\tilde{\mA} = \mA + \mI$, to allow for message passing of depth $l$ to include messages from all random walks of length $r \le l + 1$. With these definitions in hand, we define our combinatorial graph Laplacian as $\L = \mD^{-\frac{1}{2}} \tilde{\mA} \mD^{-\frac{1}{2}} $, where $\mD^{-\frac{1}{2}}$ is the diagonal degree matrix of $\tilde{\mA}$. Other normalizations are possible, but we restrict our discussion to symmetrically normalized graph Laplacians.

\textbf{Graph Signal Processing} The central challenge of signal processing on graphs is to apply the intuition from signal processing on structured grids to nonuniform and unstructured graph domains. We do so by drawing an analogy between graph and rectangular domains. The Fourier basis in rectangular domains is given by $\phi(\vx, \vk) = e^{i \vk \cdot \vx}$, where $\vk$ is the vector of frequencies, or wave numbers, and $\vx$ is the vector of positions, which are the eigenfunctions of the Laplacian in rectangular coordinates. The definition of the Fourier basis allows us to define the Fourier transform and the convolution theorem:

\begin{equation}
  f(x) = \left \{ g * h \right\} = \mathcal{F}^{-1} \left\{ \mathcal{F}(g) \cdot \mathcal{F}(h) \right\},
\end{equation}

where $\mathcal{F}$ and $\mathcal{F}^{-1}$ is the Fourier transform and inverse Fourier transform respectively; $*$ is the spatial convolution; and $\cdot$ is the point wise product. Convolutions are ubiquitous in signal processing, with spectral filtering being the application that we will focus on. From the perspective of spectral filtering, we have a signal, $h(x)$, and a filter given by $g(\cdot)$. We can either apply this filter in the spectral or conjugate spatial domain, but spectral filtering is preferred. If the optimal filter is unknown \emph{a priori}, we can learn this filter through a learning procedure. This can be done in a variety of ways, such as learning the filter in the Fourier domain. This amounts to learning a filter with infinite support and an infinite number of free parameters. While theoretically possible, this is a suboptimal technique. Instead, we frequently choose to expand into a convenient basis. In this way, we end up with:

\begin{equation}
  f(x) = \left \{ \sum_k (c_k g_k) * h \right\} = \mathcal{F}^{-1} \left\{ \sum_k c_k \mathcal{F}(g_k) \cdot \mathcal{F}(h) \right\},
\end{equation}

where $c_k$ is the expansion coefficient for each basis function given by $g_k$. Choices of basis function can include gaussians, wavelets, polynomials, or other functional forms depending on the desired properties. The implicit assumption with this filtering construction is that there are components of the signal which are spurious for a given application. Depending on the application and the signal, we might need to learn a filter that damps high or low frequency noise, or only selects a narrow set of frequencies. These are called low-pass, high-pass, and band-pass filters respectively.

Analogously, we can use the definition of the graph Laplacian to construct a Fourier basis in the graph domain. In the case of an undirected graph, the Laplacian is a symmetric positive semidefinite matrix which admits an eigensystem with orthonormal eigenvectors and real, positive, eigenvalues. These eigenvectors form the graph Fourier basis and the eigenvectors form the squared frequencies. We can then write the eigendecomposition as $\mathcal{L} = \mU^T \Lambda \mU$, where $\mU$ is the matrix of eigenvectors and $\Lambda$ is the diagonal matrix of eigenvalues.

This definition allows us to begin to define filters. The original spectral GCN paper~\cite{bruna2013spectral} learned filters of the form $g(\Theta) = \textrm{diag}(\Theta)$, where $\textrm{diag}(\theta)$ is learned set of parameters for each of the Fourier basis functions. This is then used for filtering as:

\begin{equation}
  f(x) = U \left[ g(\Theta) \cdot \left( U^T x \right) \right]
\end{equation}

While effective and highly flexible, this filtering technique has $O(n)$ learnable parameters, can be difficult to learn, and requires decomposition of $\mathcal{L}$ to construct eigenvectors $\mU$. Spectral convolutions are mathematically elegant, and the construction of the basis and learning of the filter are computationally demanding, making them not ideal in many use cases. By contrast, convolutions can also be defined in the spatial domain. Spatial convolutions are spatially localized, so they have finite size, but they aren't guaranteed to be unique which makes them computationally difficult to learn and to apply. This method makes the assumption that we can approximate $g$ as $g(\Lambda; \left\{ \Theta \right\}) = \sum_k^K \theta_k \Lambda^k$, which yields a filtered signal of the form:

\begin{equation} \label{eq:taylorexp}
  f(x) = \sum_k^K \theta_k \mathcal{L}^k x.
\end{equation}

This learned filter is more parameter efficient and spatially localized, because convolutions only extend to a distance reachable by $K$ applications of the Laplacian. Given that we desire to limit our computational cost, we will restrict ourselves to learnable functional forms that result only in matrix products because for sparse matrices, these scale as $O(| \mathcal{E} |)$. Chebyshev polynomials are one straightforward choice because they are the interpolating polynomial that provides the best function approximation under the maximum error norm. These polynomials yield a filter of the form:

\begin{equation} \label{eq:chebnet}
  g\left({\Lambda; \left\{ \Theta \right\}}\right) = \sum_k^K \theta_k T_k(\tilde{\Lambda}),
\end{equation}

where $T_k$ is the $k^{th}$ Chebyshev polynomial and $\tilde{\Lambda}$ is the eigenvalue matrix that has been rescaled to have all eigenvalues in $[-1, 1]$. The $T_k$ polynomial is given by the recurrence relationship that $T_k(x) = 2 x T_{k-1}(x) - T_{n-2}(x)$, where $T_0(x) = 1$ and $T_1(x) = x$. The rescaled $\Lambda$ is given by $2 \Lambda / \lambda_{max} - I$, where $\lambda_{max}$ is bounded from above by $\lambda_{max} \le \textrm{max}(d_i) + \textrm{max}(d_j)$, where $\textrm{max}(d_i)$ is the largest in-degree, and $d_j$ is the largest out-degree in the graph~\citep{max_eigenval}. Because these filters have their leading term as a power of $\mathcal{L}^k$, they are localized to only include vertices within $k$-hops. While these polynomials are well k-localized, we lose the frequency specificity that might enable one to construct narrow band filters. While in principal it should be possible for ChebNet to learn any form of filter, in practice they're observed to best represent wide low-pass filters~\citep{chebnet}.

\textbf{Diffusion as Convolution} We can readily observe \eqref{eq:taylorexp} and \eqref{eq:chebnet} as the Taylor and Chebyshev expansions of the matrix exponential, or heat kernel, for particular choices of $\theta$. This amounts to solving the partial differential equation;

\begin{equation}
  \frac{\partial f}{\partial t} = - c \Laplace f = - c \L f,
\end{equation}

where $\Laplace$ is the Laplace-Beltrami on the graph, and admits a solution $f(t) = \textrm{exp}(- t \L) f(0)$. Viewed in this way, we can draw the equivalence between solving the heat equation on the graph and many message passing schemes. Building on this observation, \citet{digl} observed that it is possible to define a generalized graph diffusion operator $\mS$ as $\mS = \sum^\infty_{k} \theta_k \mT^k$ where $\mT$ is a transition matrix, such as the combinatorial Laplacian. As we can see from the definition of the heat kernel, it exponentially squashes the contributions from eigenmodes with large eigenvalues, thereby providing us with a low-pass filter. Intuitively, this should help in settings where averaging out the noise from neighbors can improve performance. Following this line of thought, many graph neural network architectures can be understood as performing generalized diffusion over a graph~\citep{blend, grand}. 

From a physical perspective, the heat equation is well known to drive towards an infinitely smooth equilibrium solution; by smoothing out the initial distribution exponentially quickly in acausal ways. The heat equation is also well known to exhibit thermal bottlenecks when the heat flux and temperature gradient vectors deviate~\citep{bornoff2011heat, grossmann2000scaling}. This physical process is analogous to the oversmoothing and oversquashing problems that have plagued graph neural networks respectively. We can observe both of these physical processes in Figure~\ref{fig:diff}, which presents a comparison of heat and quantum diffusion at four different time steps. By $t=50$, we observe that the top lobe of the barbell has completely thermalized, or oversmoothed, and heat is slowly starting to leak into the center vertex, or oversquashed. In $200$ timesteps, we observe very little in the way of heat transfer. It is clear that there is a thermal bottle neck in our system, and this is hampering the flow of heat.

\section{Methodology}\label{sec:methods}

\begin{figure}[t]
  \begin{center}
    \includegraphics[width=\textwidth]{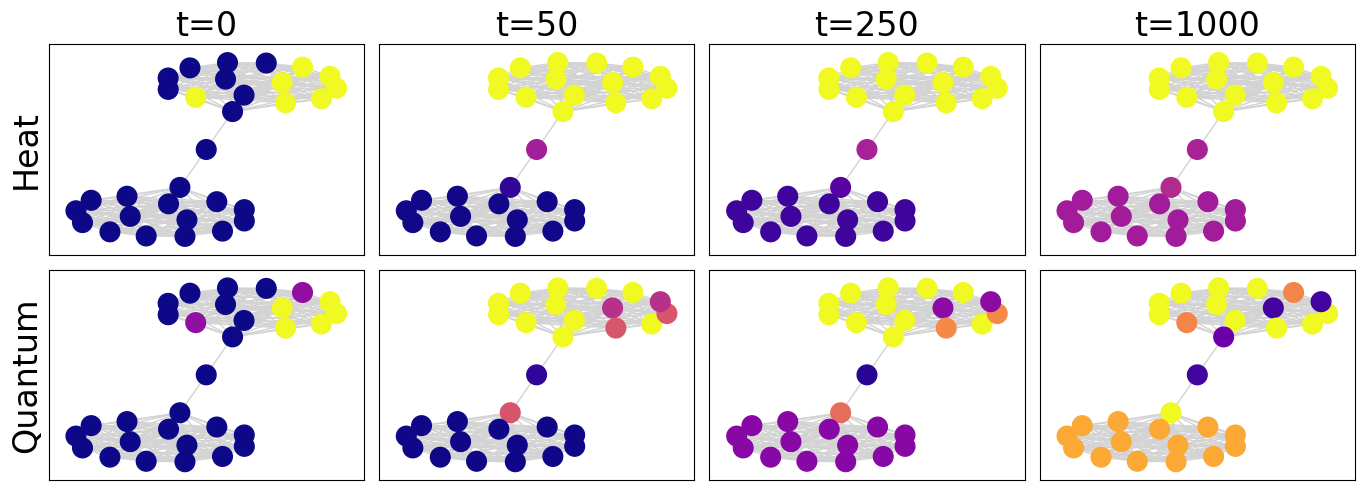}
  \end{center}
  \caption{A comparison of both heat diffusion and quantum dynamics on a barbell graph. The top row corresponds diffusion according to the heat equation of the graph, and the bottom row corresponds to propagation of the Schr\"{o}dinger equation. We simulated dynamics for 1000 unitless timesteps with the same initial distribution for both equations. We use the \texttt{plasma} colorbar, with blue and yellow corresponding to the minimum and maximum respectively. We observe the top row thermalizes within the cluster rapidly to the same temperature but encounters a bottleneck as the two ends of the barbell thermalize. By contrast, quantum dynamics exhibits oscillatory behavior both within the clusters as well as across the clusters, and probability density migrates rapidly.}
\label{fig:diff}
\end{figure}

\textbf{Quantum Convolution Kernels} Because the graph defines both the geometry of the system and its Laplaician, it is natural to ask if grounding our message passing in a different physical model would yield better results. There are many possible options including the wave equation, which propagates the sharpness of solutions; the elastic wave equation which models both forces as displacements as vector quantities; the convection equation which models the flux; the eikonal equation which simulates shortest path searches;  or Schr\"{o}dinger's equation which models quantum particles. Of these partial differential equations (PDEs), we turn our attention to Schr\"{o}dinger's equation. While structurally similar to the heat equation, its dynamics are stable, do not lead to oversmoothing, natively capture both constructive and destructive interference, and are controllable through the engineering of potential energy surfaces~\citep{chou2012adiabatic, jaffe1987time, sakurai1995modern}. Indeed, Schr\"{o}dinger's equation and the heat equation are related through a Wick rotation by $\pi/2$~\citep{popov2005imaginary}. Qualitatively, we observe in Figure~\ref{fig:diff} that unlike thermal diffusion, quantum diffusion is able to quickly pass information through the bottleneck. Later timesteps show us constructive and destructive interference that provide structure to the propagations. Videos of the propagation show oscillations across the top lobe and along the vertical axis as a function of time. Qualitatively, these dynamics do not seem to oversmooth and are less prone to oversquashing as a result. Motivated by these qualitative results, we briefly introduce Schr\"{o}dinger's equation, quantum dynamics, and quantum measurement, which are used to derive our quantum convolution kernel.

In quantum mechanics, we separate the dynamics of the underlying quantum system from the observation of physical observables. This is done because quantum wave functions, canonically denoted by $\ket{\psi(t)}$, are superpositions of many different states and $\ket{\cdot}$ is the ``ket'' using Dirac, or bra-ket, notation. This superposition ``collapses'' when we measure some detail about the configuration, because we have probabilistically projected the configuration into one of its constituent states. Dirac notation conveniently allows us to avoid the selection of basis and to ease calculations. Through the application of an observable such as position, magnetization, or energy, we collapse this wave function into a single state. The dynamics of our wave function are governed by the time dependent Schr\"{o}dinger equation, which is a parabolic partial differential equation;
\begin{equation}
  i \frac{\partial \psi(x, t)}{\partial t} = - \Laplace \psi(x, t) + V(x, t) \psi(x, t) = \mathcal{H} \psi(x, t).
\end{equation}

where $\Laplace$ is the Laplace-Beltrami operator, and $V(x, t)$ is the potential energy function. Given a configuration of the system, known as the wave function, at $t=t_1$, we can propagate it to $t_2$ by use of the unitary evolution operator, $\mathcal{U}(t_1, t_2)$ defined as $ \mathcal{U}(t_1, t_2) = e^{-i \mathcal{H} (t_2 - t_1)}$ such that $ \ket{\psi(t_2)} = \mathcal{U}(t_2, t_1) \ket{\psi(t_1)}$. Let $\mathcal{O}$ denote measurement of some observable at time $t_2$, which is performed by applying $\mathcal{O} \ket{\psi(t_2)}$. In general, an observable is a positive operator-valued measure with a spectral decomposition, \textit{i.e.} $\mathcal{O} = \sum_i \lambda_i \mathcal{P}_i$, where $\lambda_i$ is the $i^{th}$ eigenvalue of the operator and $\mathcal{P}_i$ is the projector onto the $i^{th}$ subspace. The probability of observing the state $\lambda_i$ is given by $p(\lambda_i) = \bra{\psi(t_2)} \mathcal{P}_i \ket{\psi(t_2)}$, and the expected value is given by $\bra{\psi(t_2)} \mathcal{O} \ket{\psi(t_2)}$. After observation, we have collapsed our wave function, which yields $\ket{ \bar{\psi}(t_2) } = \mathcal{P}_i \ket{\psi(t_2)}$, where the bar indicates the wavefunction has been observed. A special case of observation involves projecting our wave function into a particular representation such as the position basis, which is done by computing $\psi(x, t) = \braket{x | \psi(t)}$. This notation allows one to cleanly compute the probability that a particle at position $x_i$ is observed at $x_j$ by computing $\braket{ \psi(x_i) | \psi(x_j)} = \braket{i | j}$.

The simplest version of evolution involves the propagation of a quantum particle in some space without a potential. For such a system, the Hamiltonian is given by $\mathcal{H} = -\Laplace$, which has a similar structure to that of the heat equation. The eigenstates of $\mathcal{H}$ define a complete and orthogonal basis such at we can expand any state as;

\begin{equation}
  \ket{\psi(t)} = \sum_i c_i e^{\iu E_i t} \ket{\phi_i}
\end{equation}

where $\ket{\phi_i}$ is the $i^{th}$ eigenvector, $\iu = \sqrt{-1}$ is the imaginary unit, and $c_i$ is the expansion coefficient given by $c_i = \braket{\psi(0) | \phi_i}$. In the case of a graph, for which we have a combinatorial Laplacian, we have a finite set of eigenstates and our position operator is localized at each vertex. Therefore, the expectation value $\braket{i(0) | j(t)}$ computes the probability that a particle on vertex $i$ is observable on vertex $j$ at some time $t$. In our setting, we wish to compute the long-term steady-state distribution, and to avoid the use of complex values, so we compute the infinite time integral of the average overlap between any two vertices, and is given by;

\begin{equation}
  \int^{\infty}_0 dt \braket{i(0) | j(t)} = \sum_{\alpha, \beta} c^\dagger_{\alpha} c_\beta \phi^{\dagger}_{\alpha}(x_i) \phi_{\beta}(x_j),
\end{equation}

because $e^{\iu E_i t}$ is orthogonal for the $L^2$ norm. Because we would like to model an observation process that is frequency dependent and introduce tunable parameters into our diffusion kernel, we  return to the concept of an observation operator. It is common to engineer observation operators to, for example, match the behavior of a known piece of equipment. Using this knowledge, we engineer our measurement operator to act as a spectral filter. There are many different forms of band pass filter, but we use a Gaussian filter defined by $\mathcal{P} = \sum_i \textrm{exp}\left( - (E_i - \mu)^2 / 2 \sigma^2 \right)$, where $\mu$ and $\sigma$ are our two tunable parameters. We can use this filter to compute a filtered transition matrix given by $\braket{i(t) | P | j(t)}$. Assuming that our initial quantum state was equally delocalized across all vertices, we obtain the final expression for $\Q$:

\begin{equation} \label{eq:qdc}
  \Q(x_i, x_j) = \sum_\alpha e^{- \frac{(E_\alpha - \mu)^2}{2 \sigma^2}} \phi_\alpha^\dagger(x_i) \phi_\alpha(x_j),
\end{equation}

where $\Q$ is our Quantum Diffusion Kernel (\methodname). Intuitively, we interpret $Q(x_i, x_j)$ as the time averaged probability of transition from vertex $i$ to vertex $j$. Since we have assumed that the graph is undirected and that the $(i,j)$ matrix element is computed with a particle initially localized at $i$ and measured at $j$, the transition probabilities computed are symmetric. Analogously to GDC, we can use $\mathcal{Q}$ as our transition matrix, instead of combinatorial graph Laplacian. Doing so allows us to use \methodname with any message passing neural network by simply replacing $\mathcal{L}$ with $\mathcal{Q}$. We present a schematic of our method in Figure~\ref{fig:fig1} in which the first transformation corresponds to switching from the original graph laplacian to transition matrix.

\textbf{Sparsification} \methodname defined as $\Q(x_i, x_j)$ is a matrix $\Q_{i,j} = \Q(x_i, x_j)$, where $\Q_{i,j}$ is the probability of transition from vertex $i$ to vertex $j$. Most Graph diffusion results in a dense transition matrix, and \methodname is no different. This happens because a quantum particle starting at site $i$ will visit all vertices within its connected component given an infinite amount of time, yielding probabilities that can be small but non-zero. This is a potentially major downfall of \methodname when compared against spatial methods like Graph Diffusion Convolution~\citep{digl}. This has the potential to introduce $\mathcal{O}(N^2)$ storage costs. To address this issue, we sparsify the resulting \methodname matrix. We consider two different sparsification methods: a simple threshold based approach, or an approach that only keeps the top-k highest weighted connections. We denote the sparsified \methodname kernel as $\tilde{\Q}$. While $\Q$ was both row and column normalized, $\tilde{\Q}$ is not. Therefore, after sparsification we normalize $\tilde{Q}$ in the usual way, defining $\tilde{Q}_{sym}  = \mD^{-1/2}_{\tilde{Q}} \tilde{Q} \mD^{-1/2}_{\tilde{Q}}$. We will drop the $sym$ in the following, such that all uses of $\tilde{Q}$ are normalized.

\textbf{Efficient Diagonalization} \methodname is a spectral method, and depends on the eigendecomposition of $\L$. This is commonly viewed as too computationally demanding of a procedure because the full eigendecomposition of a matrix requires $\mathcal{O}(N^3)$ time, and the storage costs of the resulting dense eigensystem are $\mathcal{O}(N^2)$ where $N$ is the number of vertices. While this is generally true, we recognize from the form of our kernel in \eqref{eq:qdc}, we are constructing a band pass filter and are thus only interested in a subset of the eigensystem. As a result, we are able to use approximate methods that are more computationally efficient. Due to the importance of eigendecomposition to the computational sciences, this problem has received considerable attention with algorithms such as power iteration~\citep{power-iteration}, divide and conquer~\citep{divide-and-conquer}, Arnoldi iteration~\citep{arnoldi}, Lanczos iteration~\citep{lanczos, lanczos-2}, and LOBPCG~\cite{lobpcg, lobpcg-petsc}. In this work we used LOBPCG, or Locally Optimal Block Preconditioned Conjugate Gradient, because it is provides a straightforward method to compute a limited number of eigenvectors and eigenvalues while only depending on the computation of matrix vector-products. LOBPCG is also known to converge linearly, be highly numerically stable, and highly scalable -- scaling to an N of more than 144 million~\citep{lobpcg-scalability} -- making it suitable for a variety of different applications. In our applications, we use the folded spectrum method~\citep{folded-spectrum} along with LOBPCG to compute eigenvalues centered around $\mu$. If the solver is unable to converge, we retry with $\mu' = \mu + \epsilon_{\lambda}$, where $\epsilon_{\lambda} = 1e-6$. In our settings, we compute $\textrm{min}(512, N)$ eigenvalue, eigenvector pairs. Because our eigendecomposition is done once as a preprocessing step and the sparsified $\Q$ is stored, we observe no significant runtime differences for training or inference.

\textbf{Multiscale GNN} \methodname can be used as a drop-in replacement for a transition matrix for any message passing GNN. In section~\ref{sec:exp-res}, we explore using \methodname in place of $\L$ for both graph convolutional networks and graph attention networks. Because \methodname provides a band pass filter, unlike GDC which provides a low-pass filter, it is interesting to explore the message passing across both $\L$ and $\Q$ in parallel. In this setting, we pass messages in parallel using $\L$ on one side and $\Q$ on the other. We then combined messages from each tower by either adding or concatenating them together. Finally, we feed the resulting messages into a readout function for predictions. We term this method \multimethodname, because we are able to pass messages across multiple length scales of the graph.

\section{Experiments}\label{sec:exp-res}

\textbf{Experiment Setup} In an effort to ensure a fair comparison, we optimized the hyper-parameters of all models on all data sets. We performed 250 steps of hyper-parameter optimization for each method, and the hyper-parameter search was performed using \textsc{Optuna}, a popular hyper-parameter optimization framework. All tuning was performed on the validation set, and we report the test-results associated with the hyper-parameter settings that maximize the validation accuracy. The parameters, and the distributions from which they were drawn, are reported in Appendix~\ref{appendix:model-details}. All experiments were run using \textsc{PyTorch Geometric} 2.3.1 and \textsc{PyTorch} 1.13, and all computations were run on an \texttt{Nvidia DGX A100} machine with 128 \texttt{AMD Rome 7742} cores and 8 \texttt{Nvidia A100} GPUs.

Because our method can be viewed as a Laplacian preprocessing technique, we use \methodname in place of the traditional Laplacian in both graph convolution networks (GCN)~\citep{gcn, zhang2019graph} and graph attention networks (GAT)~\citep{gat}. \methodname is similar in structure to graph diffusion convolution (GDC)~\citep{digl} and SDRL~\citep{topping2021understanding}, so we have chosen to compare \methodname to both GDC and SDRL in addition to an unprocessed Laplacian in both a GCN and a GAT.

\textbf{Datasets} We evaluated our method on 9 data sets: \textsc{Cornell}, \textsc{Texas}, and \textsc{Wisconsin} from the WebKB dataset; \textsc{Chameleon} and \textsc{Squirrel} from the Wiki dataset; Actor from the film dataset; and citation graphs \textsc{Cora}, \textsc{Citeseer}, and \textsc{Pubmed}. Where applicable, we use the same data splits as \citet{geomgcn}. Results are then averaged over all splits, and the average and standard deviation are reported. These datasets represent a mix of standard heterophilic and homophilic graph datasets. The statistics for the datasets are presented in the first three rows of Table~\ref{tab:bigtable}, where we have used the definition proposed by \citet{geomgcn} for homophily, given by:

\begin{equation}
  \mathcal{H}(\gG) = \frac{1}{\left|\mathcal{V}\right|} \sum_{v \in \mathcal{V}} \sum_{u \in \mathcal{N}_v} \frac{ \1_{l(v) = l(u)} }{|\mathcal{N}_v|}
\end{equation}

where $\mathcal{N}$ is the neighborhood operator and $l$ is the operator that returns the label of the vertex.

\textbf{Node Classification} We present the results from our experiments in Table~\ref{tab:bigtable}. We observe that \methodname provides improvements in accuracy across the heterophilic datasets, but seems to provide mixed results for Cora, Citeseer, and Pubmed. By using \multimethodname, we see that multi-scale modeling appears to provide improvements across all datasets. This validates our hypothesis that \methodname can provide a viable step forward to improving GNN performance.

\begin{table}[t]
  \caption{Dataset statistics and experimental results on common node classification benchmarks. $\mathcal{H}$, $|\mathcal{V}|$,  $|\mathcal{E}|$ denote degree of homophily, number of vertices and  number of edges, respectively. Top results for each of the GCN and GAT families are highlighted in bold.}
  \label{tab:bigtable}
  \begin{center}
\resizebox{\textwidth}{!}{
  \begin{tabular}{lccccccccc}
\Xhline{2\arrayrulewidth}
                         & Cornell                   & Texas                     & Wisconsin                   & Chameleon                 & Squirrel                  & Actor                     & Cora                      & Citeseer                  & Pubmed \\
  \hline
   $\mathcal{H}$         &   0.11                    &  0.06                     &  0.16                       &  0.25                     &   0.22                    &  0.24                     & 0.83                      &   0.71                    &  0.79           \\
  $|\mathcal{V}|$        &   183                     &  183                      &  251                        &  2,277                    &   5,201                   &  7,600                    & 2,708                     &   3,327                   &  18,717         \\
  $|\mathcal{E}|$        &   280                     &  295                      &  466                        &  31,421                   &   198,493                 &  26,752                   & 5,278                     &   4,676                   &  44,327         \\
  \hline
    GCN                  & 45.68 $\pm$ 7.30          & 63.51 $\pm$ 5.70          & 59.22 $\pm$ 4.28            & 41.16 $\pm$ 1.71          & 27.89 $\pm$ 1.21          & 29.32 $\pm$ 1.07          & 87.46 $\pm$ 1.11          & 76.61 $\pm$ 1.28          & \textbf{88.47 $\pm$ 0.39} \\
    GCN+GDC              & 47.03 $\pm$ 5.69          & 63.51 $\pm$ 6.07          & 57.25 $\pm$ 2.88            & 40.42 $\pm$ 2.93          & 27.97 $\pm$ 0.93          & 29.14 $\pm$ 0.91          & 87.63 $\pm$ 0.91          & 76.58 $\pm$ 1.21          & 88.46 $\pm$ 0.55          \\
    GCN+SDRL             & 45.14 $\pm$ 8.20          & 62.97 $\pm$ 5.55          & 57.84 $\pm$ 1.52            & 40.55 $\pm$ 1.52          & 28.17 $\pm$ 0.97          & 29.07 $\pm$ 1.03          & 87.44 $\pm$ 1.10          & \textbf{76.85 $\pm$ 1.47} & \textbf{88.47 $\pm$ 0.34} \\
    GCN+BPDC             & 60.81 $\pm$ 5.95          & 68.92 $\pm$ 6.54          & 63.73 $\pm$ 5.28            & 50.44 $\pm$ 1.77          & 40.37 $\pm$ 1.17          & 31.46 $\pm$ 1.04          & 85.86 $\pm$ 1.17          & 74.70 $\pm$ 1.34          & 84.55 $\pm$ 0.56          \\
    GCN+QDC              & 63.78 $\pm$ 9.68          & 72.70 $\pm$ 6.67          & \textbf{65.29 $\pm$ 6.80}   & \textbf{53.22 $\pm$ 1.56} & 40.62 $\pm$ 1.94          & \textbf{35.08 $\pm$ 0.64} & 86.00 $\pm$ 1.56          & 75.10 $\pm$ 1.52          & 84.65 $\pm$ 0.44          \\
    GCN+\multimethodname & \textbf{66.22 $\pm$ 5.44} & \textbf{73.78 $\pm$ 4.53} & 64.71 $\pm$ 4.47            & 54.71 $\pm$ 2.79          & \textbf{42.24 $\pm$ 1.73} & 30.55 $\pm$ 1.45          & \textbf{87.85 $\pm$ 0.85} & 76.72 $\pm$ 1.49          & 88.32 $\pm$ 0.47          \\
  \hline
    GAT                  & 60.81 $\pm$ 8.40          & 68.11 $\pm$ 5.24          & 63.14 $\pm$ 7.58            & 44.89 $\pm$ 1.64          & 31.47 $\pm$ 1.44          & 30.48 $\pm$ 1.17          & 86.68 $\pm$ 1.64          & 75.64 $\pm$ 1.55          & 84.11 $\pm$ 0.70          \\
    GAT+GDC              & 61.89 $\pm$ 7.30          & 68.11 $\pm$ 5.09          & 63.33 $\pm$ 3.62            & 45.96 $\pm$ 1.94          & 31.66 $\pm$ 1.72          & 31.18 $\pm$ 0.76          & 86.46 $\pm$ 1.20          & 75.92 $\pm$ 1.10          & 87.53 $\pm$ 0.55          \\
    GAT+SDRL             & 59.19 $\pm$ 6.33          & 67.30 $\pm$ 4.90          & 63.92 $\pm$ 5.20            & 45.88 $\pm$ 1.93          & 31.76 $\pm$ 1.00          & 31.13 $\pm$ 0.76          & 85.29 $\pm$ 1.34          & 75.90 $\pm$ 1.27          & 87.47 $\pm$ 0.48          \\
    GAT+QDC              & \textbf{77.57 $\pm$ 3.83} & \textbf{87.57 $\pm$ 5.56} & \textbf{88.04 $\pm$ 3.33}   & 50.90 $\pm$ 2.16          & 35.38 $\pm$ 1.81          & 35.57 $\pm$ 1.05          & 84.68 $\pm$ 1.54          & 75.21 $\pm$ 1.30          & 87.55 $\pm$ 0.31          \\
    GAT+\multimethodname & 77.03 $\pm$ 4.05          & 86.22 $\pm$ 5.60          & \textbf{88.04 $\pm$ 4.06}   & \textbf{52.08 $\pm$ 2.60} & \textbf{36.90 $\pm$ 1.11} & \textbf{36.55 $\pm$ 1.22} & \textbf{87.73 $\pm$ 0.74} & \textbf{76.39 $\pm$ 1.32} & \textbf{87.59 $\pm$ 0.38} \\
\Xhline{2\arrayrulewidth}
  \end{tabular}
  }
\end{center}
\end{table}

\begin{figure}[t]
  \begin{center}
    \includegraphics[width=\textwidth]{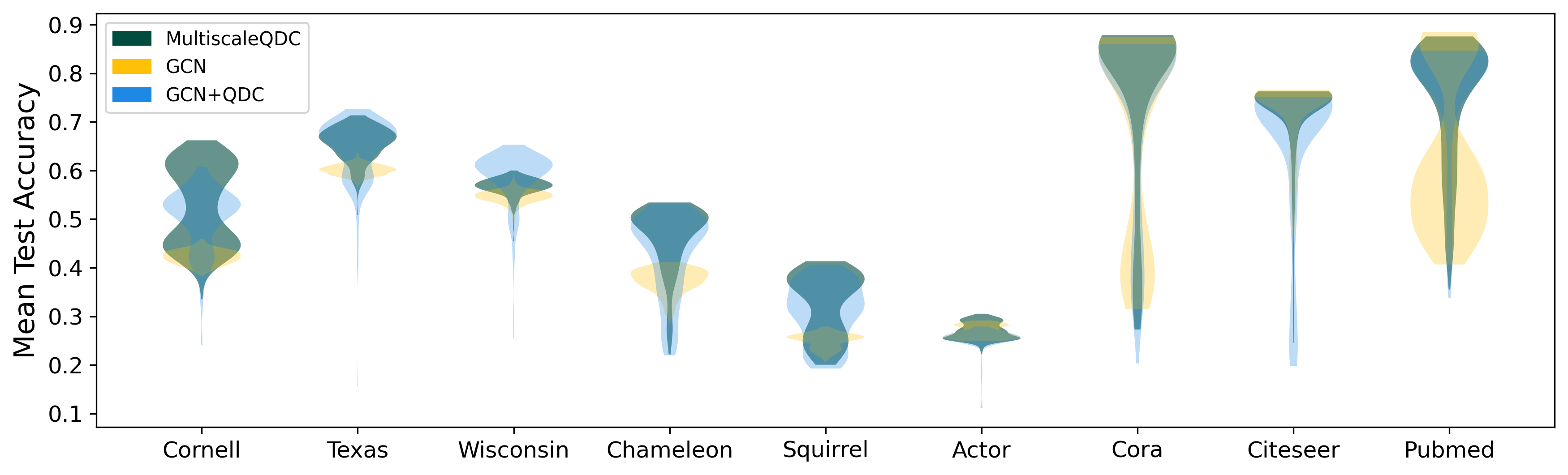}
  \end{center}
  \caption{Violin plots of our experiments GCN+\multimethodname (green), GCN (yellow), and GCN+\methodname (blue), where these plots are generated by aggregating over all experiments associated with each model. We observe that both GCN+\methodname and GCN+\multimethodname generally have a high density of near-optimal configurations.}
  \label{fig:violin}
\end{figure}

\textbf{Analysis of Hyper-parameters} \methodname has multiple hyperparameters that we tuned as part of our experiments. To understand the sensitivity of our method to these hyperparameters, we first present a violin plot in Figure~\ref{fig:violin}, which plots a kernel density estimate of the model performances from the experiments on a GCN, GCN+\methodname, and \multimethodname. In the case of the Cornell dataset, we clearly observe that \multimethodname has two humps, which correspond to the GCN and \methodname distributions. We see similar patterns in the Texas, Wisconsin, Squirrel, and Actor datasets as well. Furthermore, we clearly see that there are many experimental settings that out-perform the baseline model. 

\begin{figure}[t]
  \begin{center}
    \includegraphics[width=\textwidth]{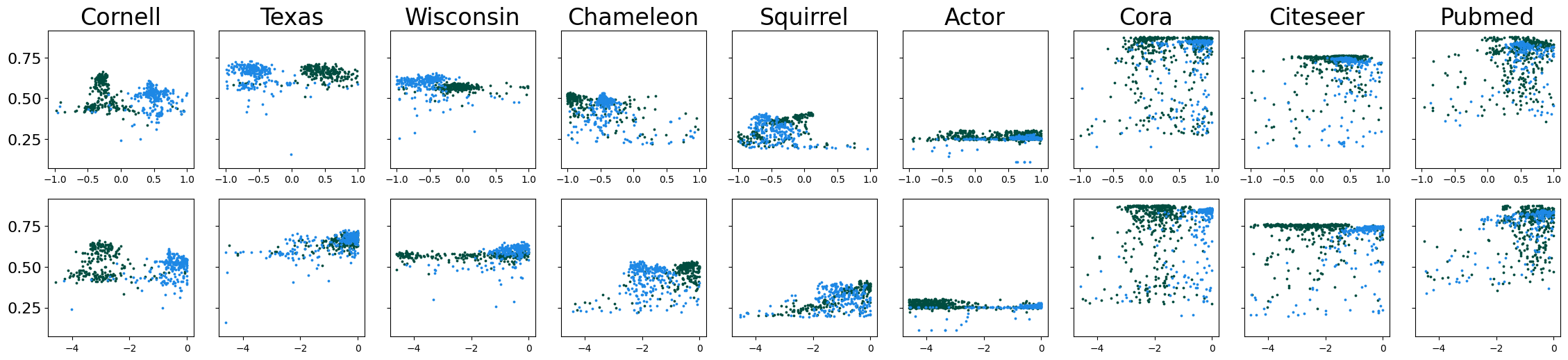}
  \end{center}
  \caption{Scatter plots of mean test accuracy plotted against hyperparameters $\mu$ and $\ln(\sigma)$ in the first and second rows respectively for GCN+\methodname (blue) and GCN+\multimethodname (green). We observe that each of \methodname and \multimethodname are robust with respect to deviations in each of the hyperparameters.}
  \label{fig:scatter}
\end{figure}

We next turn our attention to the sensitivity of our model to $\mu$ and $\sigma$ for both \methodname and \multimethodname models by plotting mean test accuracy against $\mu$ and $\sigma$ in the first and second rows of Figure~\ref{fig:scatter} respectively. We have plotted both GCN+\methodname (blue) and our \multimethodname (green) on the same plot. We observe that in general, there are many settings of $\mu$ and $\sigma$ that provide near equivalent performance which indicates that our method is robust to potentially suboptimal choice of hyperparameters. Interestingly, we find that the optimal $\mu$s for GCN+\methodname and our \multimethodname model are quite different. This is because in the \multimethodname case, we are looking for eigenvectors that correct for any deficiencies in the original combinatorial Laplacian.

\textbf{Importance of Quantum Dynamics} \methodname has two components, the quantum dynamics and the choice of filter. In our development of our method we have chosen to use a Gaussian filter because it models inhomogeneous broadening, which is a physical effect that is observed as a transition frequencies taking on a Gaussian profile due to microscopic details of the system such as atomic motion. This physical model is intuitively sensible if we imagine that our vertices are analogous to the atoms, and the latent embeddings are the atomic positions. While we can provide physical arguments from analogy for this choice, citation networks are not molecular systems. This raises the question of whether the Gaussian form of our filter is important, or whether any band-pass filter would be sufficient. To answer this question we implemented a variant of \methodname given by

\begin{equation}
  \mathcal{B}(x_i, x_j) = \sum_\alpha \sigma\left( E_\alpha - \mu + \gamma \right) \sigma\left( \mu + \gamma - E_\alpha \right)  \phi_\alpha^\dagger(x_i) \phi_\alpha(x_j),
\end{equation}

where $\sigma(\cdot)$ is the logistic sigmoid function, $\mu$ is the center of our bandpass filter, $\gamma$ is the width of our band-pass filter, and $\mathcal{B}$ is the band-pass version of \methodname which we term the Band Pass Diffusion Convolution(BPDC). Using this filter, we performed experiments on a range of data sets using BPDC as our transition matrix with a GCN and have presented those results below in Table~\ref{tab:bigtable}. We observe that BPDC is able to provide significant lift across the heterophilic datasets, but that lift is in general smaller than that observed with \methodname.

\begin{figure}[t]
  \begin{center}
    \includegraphics[width=0.8 \textwidth]{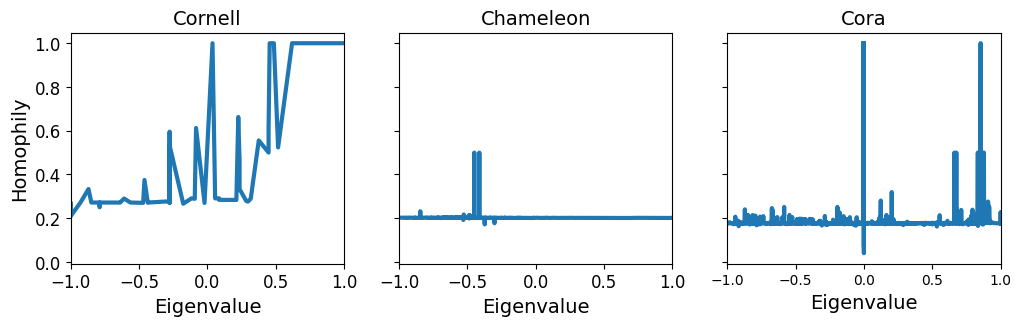}
  \end{center}
  \caption{Plots of the homophily as a function of the eigenvalues. We observe that homophily has a strong spectral dependence, and that the mid-band peaks in homophily agree with recovered optimal $\mu$s.}
  \label{fig:homophily_dep}
\end{figure}

\textbf{Spectral Dependence of Homophily} It has previously been observed that the performance of Graph Convolution models correlates with the homophily of the graph, which motivates us to ask whether homophily is spectrally dependent. To answer this question, we constructed adjacency matricies from subsets of the eigenvectors that corresponded to each unique eigenvalue. In the case where the eigenvalues were degenerate, we computed the mean homophily. We then sparsified the resulting adjacency matrix by removing all entries smaller than $1e-7$, and plotted the results in Figure~\ref{fig:homophily_dep}. We observe that the homophily is highly spectrally dependent. In the case of the \texttt{Cornell} dataset, we observe that the dataset is generally quite heterophilic but becomes more homophilic in higher portions of the spectrum; and observe that the $\mu$ cluster for GCN+\methodname in Figure~\ref{fig:scatter} corresponds to this highly homophilic region. We observe spectral variations of homophily for \texttt{Chameleon} and \texttt{Cora} as well; and note the same agreement between the optimal $\mu$ and this observed spectral peaks in the homophily curves. 

% \textbf{Oversquashing} Finally, we turn our attention to the question of oversquashing and whether \methodname

\section{Conclusion}
In this work we have introduced a quantum diffusion kernel that we have termed \methodname, and a multiscale model that we have termed \multimethodname. We have motivated this convolution kernel through a deep connection with quantum dynamics on graphs. In experiments we have shown that \methodname generally helps in cases of heterophilic node classification, and \multimethodname seems to improve in both homophilic and heterophilic node classifications settings.

While we are able to use iterative, matrix-free, eigensolvers, our method is still more expensive than spatial convolutions. Additionally, propagating gradients through approximate eigensolvers is quite challenging, making it difficult to optimize the parameters of the diffusion kernel during training time. Finally, because our method is spectral, we are only able to use this method in transductive settings. We believe that quantum convolution in the spatial domain will open up avenues to address these issues, and are excited to explore this approach in followup work.

\bibliography{paper}
\bibliographystyle{tmlr}

\newpage
\appendix
\section{Appendix}

\subsection{Model Details}\label{appendix:model-details}

We performed 250 steps of hyper-parameter optimization for each of the models presented in Table~\ref{tab:bigtable}. All training runs were run with a maximum of 1000 steps for each split, with early stopping turned on after 50 steps. In the interest of reproducibility, we outline the parameters and ranges that we're optimized for each model below.

%%%%%%%%%%%%%%%%%%%%%%%%%%%%% GCN

\begin{table}[t]
  \caption{Hyper-parameter ranges that we optimized over for our GCN.}
  \label{tab:gcn-params}
  \begin{center}
    \begin{tabular}{lcc}
\Xhline{2\arrayrulewidth}
  Parameters & Distribution & Values \\
\hline
  Number of Layers & Categorical & [1, 2] \\  
  Hidden Dim Size & Categorical & [2, 4, 8, 16, 32, 64, 128] \\  
  Dropout percentage & Uniform & [0, 0.99] \\
  Learning Rate & Loguniform & [1e-4, 1e-1] \\  
  Weight Decay & uniform & [0.0, 0.9] \\  
\Xhline{2\arrayrulewidth}
    \end{tabular}
  \end{center}
\end{table}

\begin{table}[t]
  \caption{Hyper-parameter ranges that we optimized over for our GCN+GDC.}
  \label{tab:gcn-gdc-params}
  \begin{center}
    \begin{tabular}{lcc}
\Xhline{2\arrayrulewidth}
  Parameters & Distribution & Values \\
\hline
  Number of Layers & Categorical & [1, 2] \\  
  Hidden Dim Size & Categorical & [2, 4, 8, 16, 32, 64, 128] \\  
  Dropout percentage & Uniform & [0, 0.99] \\
  GDC-$\alpha$ & uniform & [0.001, 0.5] \\  
  GDC-$\epsilon$ & uniform & [1e-7, 1e-1] \\  
  Learning Rate & Loguniform & [1e-4, 1e-1] \\  
  Weight Decay & uniform & [0.0, 0.9] \\  
\Xhline{2\arrayrulewidth}
    \end{tabular}
  \end{center}
\end{table}

\begin{table}[t]
  \caption{Hyper-parameter ranges that we optimized over for our GCN+\methodname.}
  \label{tab:gcn-qdc-params}
  \begin{center}
    \begin{tabular}{lcc}
\Xhline{2\arrayrulewidth}
  Parameters & Distribution & Values \\
\hline
  Number of Layers & Categorical & [1, 2] \\  
  Hidden Dim Size & Categorical & [2, 4, 8, 16, 32, 64, 128] \\  
  Dropout percentage & Uniform & [0, 0.99] \\
  \methodname-$\mu$ & uniform & [-1, 1] \\  
  \methodname-$\sigma$ & uniform & [0.1, 1.0] \\  
  \methodname-$\epsilon$ & loguniform & [1e-7, 1e-1] \\  
  Learning Rate & Loguniform & [1e-4, 1e-1] \\  
  Weight Decay & uniform & [0.0, 0.9] \\  
\Xhline{2\arrayrulewidth}
    \end{tabular}
  \end{center}
\end{table}

\begin{table}[t]
  \caption{Hyper-parameter ranges that we optimized over for our \multimethodname.}
  \label{tab:gcn-multiscaleqdc-params}
  \begin{center}
    \begin{tabular}{lcc}
\Xhline{2\arrayrulewidth}
  Parameters & Distribution & Values \\
\hline
  GCN Number of Layers & Categorical & [1, 2] \\  
  GCN Hidden Dim Size & Categorical & [2, 4, 8, 16, 32, 64, 128] \\  
  GCN Dropout percentage & Uniform & [0, 0.99] \\

  \methodname Number of Layers & Categorical & [1, 2] \\  
  \methodname Hidden Dim Size & Categorical & [2, 4, 8, 16, 32, 64, 128] \\  
  \methodname Dropout percentage & Uniform & [0, 0.99] \\

  \methodname-$\mu$ & Uniform & [-1, 1] \\  
  \methodname-$\sigma$ & Uniform & [0.1, 1.0] \\  
  \methodname-$\epsilon$ & loguniform & [1e-7, 1e-1] \\  
  Combinator & Categorical & [ concat, add] \\
  Learning Rate & Loguniform & [1e-4, 1e-1] \\  
  Weight Decay & Uniform & [0.0, 0.9] \\  
\Xhline{2\arrayrulewidth}
    \end{tabular}
  \end{center}
\end{table}

%%%%%%%%%%%%%%%%%%%%%%%%%%%%%%%%%% GAT

\begin{table}[t]
  \caption{Hyper-parameter ranges that we optimized over for our GAT.}
  \label{tab:gat-params}
  \begin{center}
    \begin{tabular}{lcc}
\Xhline{2\arrayrulewidth}
  Parameters & Distribution & Values \\
\hline
  Number of Layers & Categorical & [1, 2] \\  
  Hidden Dim Size & Categorical & [2, 4, 8, 16, 32, 64, 128] \\  
  Number of Heads & Categorical & [1, 2, 3, 4, 5] \\  
  Dropout percentage & Uniform & [0, 0.99] \\
  Learning Rate & Loguniform & [1e-4, 1e-1] \\  
  Weight Decay & Uniform & [0.0, 0.9] \\  
\Xhline{2\arrayrulewidth}
    \end{tabular}
  \end{center}
\end{table}

\begin{table}[t]
  \caption{Hyper-parameter ranges that we optimized over for our GAT+GDC.}
  \label{tab:gat-gdc-params}
  \begin{center}
    \begin{tabular}{lcc}
\Xhline{2\arrayrulewidth}
  Parameters & Distribution & Values \\
\hline
  Number of Layers & Categorical & [1, 2] \\  
  Hidden Dim Size & Categorical & [2, 4, 8, 16, 32, 64, 128] \\  
  Number of Heads & Categorical & [1, 2, 3, 4, 5] \\  
  Dropout percentage & Uniform & [0, 0.99] \\
  GDC-$\alpha$ & Uniform & [0.001, 0.5] \\  
  GDC-$\epsilon$ & Uniform & [1e-7, 1e-1] \\  
  Learning Rate & Loguniform & [1e-4, 1e-1] \\  
  Weight Decay & Uniform & [0.0, 0.9] \\
\Xhline{2\arrayrulewidth}
    \end{tabular}
  \end{center}
\end{table}

\begin{table}[t]
  \caption{Hyper-parameter ranges that we optimized over for our GAT+\methodname.}
  \label{tab:gat-qdc-params}
  \begin{center}
    \begin{tabular}{lcc}
\Xhline{2\arrayrulewidth}
  Parameters & Distribution & Values \\
\hline
  Number of Layers & Categorical & [1, 2] \\  
  Hidden Dim Size & Categorical & [2, 4, 8, 16, 32, 64, 128] \\  
  Number of Heads & Categorical & [1, 2, 3, 4, 5] \\  
  Dropout percentage & Uniform & [0, 0.99] \\
  \methodname-$\mu$ & Uniform & [-1, 1] \\  
  \methodname-$\sigma$ & Uniform & [0.1, 1.0] \\  
  \methodname-$\epsilon$ & loguniform & [1e-7, 1e-1] \\  
  Learning Rate & Loguniform & [1e-4, 1e-1] \\  
  Weight Decay & Uniform & [0.0, 0.9] \\
\Xhline{2\arrayrulewidth}
    \end{tabular}
  \end{center}
\end{table}

\begin{table}[t]
  \caption{Hyper-parameter ranges that we optimized over for our Multiscale GAT+\methodname.}
  \label{tab:gat-multiscale-qdc-params}
  \begin{center}
    \begin{tabular}{lcc}
\Xhline{2\arrayrulewidth}
  Parameters & Distribution & Values \\
\hline
  GAT Number of Layers & Categorical & [1, 2] \\  
  GAT Hidden Dim Size & Categorical & [2, 4, 8, 16, 32, 64, 128] \\  
  GAT Number of Heads & Categorical & [1, 2, 3, 4, 5] \\  
  GAT Dropout percentage & Uniform & [0, 0.99] \\

  \methodname Number of Layers & Categorical & [1, 2] \\  
  \methodname Hidden Dim Size & Categorical & [2, 4, 8, 16, 32, 64, 128] \\  
  \methodname Number of Heads & Categorical & [1, 2, 3, 4, 5] \\  
  \methodname Dropout percentage & Uniform & [0, 0.99] \\
  \methodname-$\mu$ & Uniform & [-1, 1] \\  
  \methodname-$\sigma$ & Uniform & [0.1, 1.0] \\  
  \methodname-$\epsilon$ & loguniform & [1e-7, 1e-1] \\  

  Combinator & Categorical & [ concat, add] \\
  Learning Rate & Loguniform & [1e-4, 1e-1] \\  
  Weight Decay & Uniform & [0.0, 0.9] \\
\Xhline{2\arrayrulewidth}
    \end{tabular}
  \end{center}
\end{table}

\end{document}